\documentclass[runningheads]{llncs}
\usepackage[T1]{fontenc}
\usepackage{graphicx,verbatim}
\usepackage{booktabs}
\usepackage{multirow}
\usepackage{tikz}
\usepackage{url}
\begin{document}
\title{Causal-Adversarial Probing of Clinical Covariates for Prostate MRI Grading}
\titlerunning{Causal-Adversarial Probing}
%
\author{Yipei Wang\inst{1,2}\orcidID{0000-0002-9589-7177} \and
Shiqi Huang\inst{1,2}\orcidID{0000-0003-1348-3817} \and
Wen Yan\inst{1,2}\orcidID{0000-0002-3962-5994} \and
Weixi Yi\inst{1,2}\orcidID{0009-0002-9889-9853} \and
Dean C. Barratt\inst{1,2}\orcidID{0000-0003-2916-655X} \and
Mark Emberton\inst{3,4}\orcidID{0000-0003-4230-0338} \and
Daniel C. Alexander\inst{1,5}\orcidID{0000-0003-2439-350X} \and
Veeru Kasivisvanathan\inst{3,4,6,7}\orcidID{0000-0002-0832-382X} \and
Yipeng Hu\inst{1,2}\orcidID{0000-0003-4902-0486} 
}
\authorrunning{Y. Wang et al.}
%
\institute{UCL Hawkes Institute, University College London, London, UK \\ \email{yipei.wang@ucl.ac.uk} \and
Department of Medical Physics and Biomedical
Engineering, University College London, London, UK \\ \and
Department of Urology, University College London Hospital, London, UK \\ \and 
Division of Surgery and Interventional Science, University College London, London, UK \\ \and
Department of Computer Science, University College London, London, UK \\ \and
Centre for Urology Imaging, Prostate, AI and Surgical Studies (COMPASS) Research Group, Division of Surgery and Interventional Science, University College London, London, UK \\ \and
Department of Urology, Comprehensive Cancer Center, Medical University of Vienna, Vienna, Austria 
}
  
\maketitle              
\begin{abstract}

Deep learning models for prostate MRI-based cancer grading may encode clinical covariates that either reflect useful disease-related signal or non-generalising shortcut information, but their role is usually assumed. We propose a causal-reasoning framework for probing covariate dependence in MRI-based International Society of Urological Pathology (ISUP) Grade Group prediction. Rather than treating mpMRI as a direct cause of grade, we model MRI appearance and ISUP grade as observations of latent tumour pathology, and test whether candidate clinical variables act as nuisance correlates, disease-related proxies, or irrelevant covariates in the learned representation. We implement this using an adversarial framework that suppresses the decodability of individual clinical covariate at a time while preserving MRI-based grade prediction. The approach is developed and evaluated on 2,903 prostate MRI examinations, with external validation on 576 patients. We report a set of interesting and previously under-explored imaging-to-clinical-variable interactions in the context of deep learning generalisation. For examples, in binary ISUP Grade Group $\geq$2 classification, suppressing age, BMI, and alcohol use improved AUC by 1.23\%, 0.84\%, and 1.42\%, respectively (all $p<0.05$), suggesting reduced non-generalising covariate information; In contrast, suppressing PSA and prostate volume degraded AUC by 1.91\% and 7.61\% (all $p<0.001$), indicating that these variables carried task-relevant signal. These findings show that adversarial covariate suppression can provide a practical representation-level analysis for distinguishing potentially harmful dependence from informative signal in prostate MRI grading models.


\end{abstract}
\section{Introduction}
Multiparametric MRI (mpMRI) is increasingly important in the diagnostic pathway for men with suspected prostate cancer (PCa), with prospective study evidence showing that MRI-informed biopsy strategies detect more clinically significant cancer while reducing the over-diagnosis of clinically insignificant disease and the number of unnecessary biopsies \cite{ahmed2017diagnostic,kasivisvanathan2018mri}. Nevertheless, definitive risk stratification still relies on histopathological grading based on the International Society of Urological Pathology (ISUP) scale \cite{epstein20162014}, derived from histopathological assessment of biopsy or prostatectomy tissue.
Since MRI is non-invasive and provides whole-gland anatomical and functional information, there has been a growing interest in predicting clinically significant disease and estimating cancer risk that is at least as indicative as ISUP Grade Group directly from prostate MRI. Prior studies have shown that deep learning can support clinically significant cancer detection and MRI-based aggressiveness assessment, including direct prediction of Gleason score or ISUP grade from prostate MRI \cite{schelb2019classification,saha2024artificial,cao2019joint,pellicer2022deep,chaddad2020deep,xu2024poisson}.


Most MRI-based grading studies are established around a single data-driven objective: maximising predictive accuracy. Clinical and demographic variables are therefore typically treated as additional inputs to improve prediction or as cohort descriptors to be balanced or stratified. Much less attention has been paid to a more fundamental question: when a network learns the association between MRI appearance and ISUP grade, does its representation encode a candidate clinical covariate because it is a potential confounder, a disease-related proxy, or an irrelevant correlate? Distinguishing between these roles is important for understanding whether such a covariate should be retained as useful signal that carries disease knowledge or reduced as a potential source of non-generalisable shortcut learning which may need to be suppressed from the learned representation. We therefore frame each clinical variable as a candidate covariate whose role in the learned MRI representation should be tested rather than assumed.

In general, medical imaging models are vulnerable to shortcut learning because they are trained on, often limited in size, observational data, where labels, acquisition protocols, sites, and patient characteristics are often correlated due to factors that are not generalisable. Prior studies have shown that radiology models can exploit hospital-, scanner-, or cohort-specific signals rather than pathology-specific features, leading to poor generalisation across settings \cite{zech2018variable,degrave2021ai,geirhos2020shortcut}. Causal machine learning provides technical framework for separating causal or generative relations from accidental, non-causal correlations and has motivated methods for harmonisation, confounder-invariant learning, and causal intervention in medical imaging \cite{scholkopf2022causality,wang2021harmonization,zhao2020training,zhang2025causalmixnet}. 
However, practical approaches for identifying and exploiting potential causal clinical relationships remain limited, and have not been systematically demonstrated between MRI-based ISUP grade prediction and specific clinical variables in machine learning applications.

Motivated by this gap, we use causal reasoning in this study to discover candidate covariates whose information is encoded in MRI representations and whose suppression changes ISUP prediction. As illustrated in Fig.~\ref{fig:causal_graph}, we do not assume that MRI causally determines ISUP grade. Instead, MRI appearance and ISUP grade are treated as complementary observations of latent tumour pathology, while a candidate covariate may form a potential back-door \cite{mcdonald2002judea}, share a cause with MRI or grade through latent pathology, or have little relation to the prediction task. Our approach does not prove causality or perform conventional back-door adjustment by subgroup comparison or conditioning, which would require carefully predefine variable controlling configuration, additional model training and much more data for fair comparison between these models. Instead, we propose to perform a representation-level sensitivity analysis: if reducing the decodability of a covariate from the MRI representation (e.g. for predicting ISUP grading) shifts grading performance, this learning performance shift indicates a possible causality role for the covariate in the learned predictor.
 
We implement this hypothesis with an adversarial invariance framework. Given an encoder trained for ISUP prediction, a covariate head attempts to recover a selected clinical variable of interest from the learned representation, while the encoder is trained to preserve grading information and reduce the decodability of that covariate, akin to adversarial domain adaptation \cite{ganin2016domain,tzeng2017adversarial}, in both cases representation is incentivised to ``ignore'' this clinical variable. Repeating this procedure for each variable allows us to test whether suppressing each covariate improves, worsens, or leaves unchanged the MRI-only grading model. 
Our contributions are: (1) a causal formulation of MRI-based ISUP grade prediction that distinguishes candidate confounders, disease-related proxies, and irrelevant covariates;
(2) a gradient-reversal adversarial framework for variable-wise covariate suppression without requiring clinical variables at inference; 
and (3) an empirical analysis on $2,614$ patients from a single referral centre with a diverse demographic population, demonstrating heterogeneous performance shifts after suppressing different clinical covariates.
Code implementation is available at \url{https://github.com/pipiwang/CausalProbing}.

\section{Method}\label{sec:method}
\subsection{Causal formulation of clinical-variable effects}

Let \(X\) denote prostate MRI, \(Y\) denote ISUP Grade Group, and \(C\) denote one commonly available clinical variable, in this study, considered a candidate covariate. \(U\) denotes unobserved tumour pathology, 
which may generate both MRI appearance and pathological grade. Under this formulation, MRI is not treated as a direct cause of grade. Instead, \(X\) and \(Y\) are observed manifestations of the latent disease process, represented conceptually by \(U\rightarrow X\) and \(U\rightarrow Y\). The problem is therefore to extract grade-relevant information about \(U\) from \(X\), while evaluating if information about \(C\) is encoded in the learned representation.

\begin{figure}[htb]
    \centering
    \includegraphics[width=0.9\textwidth]{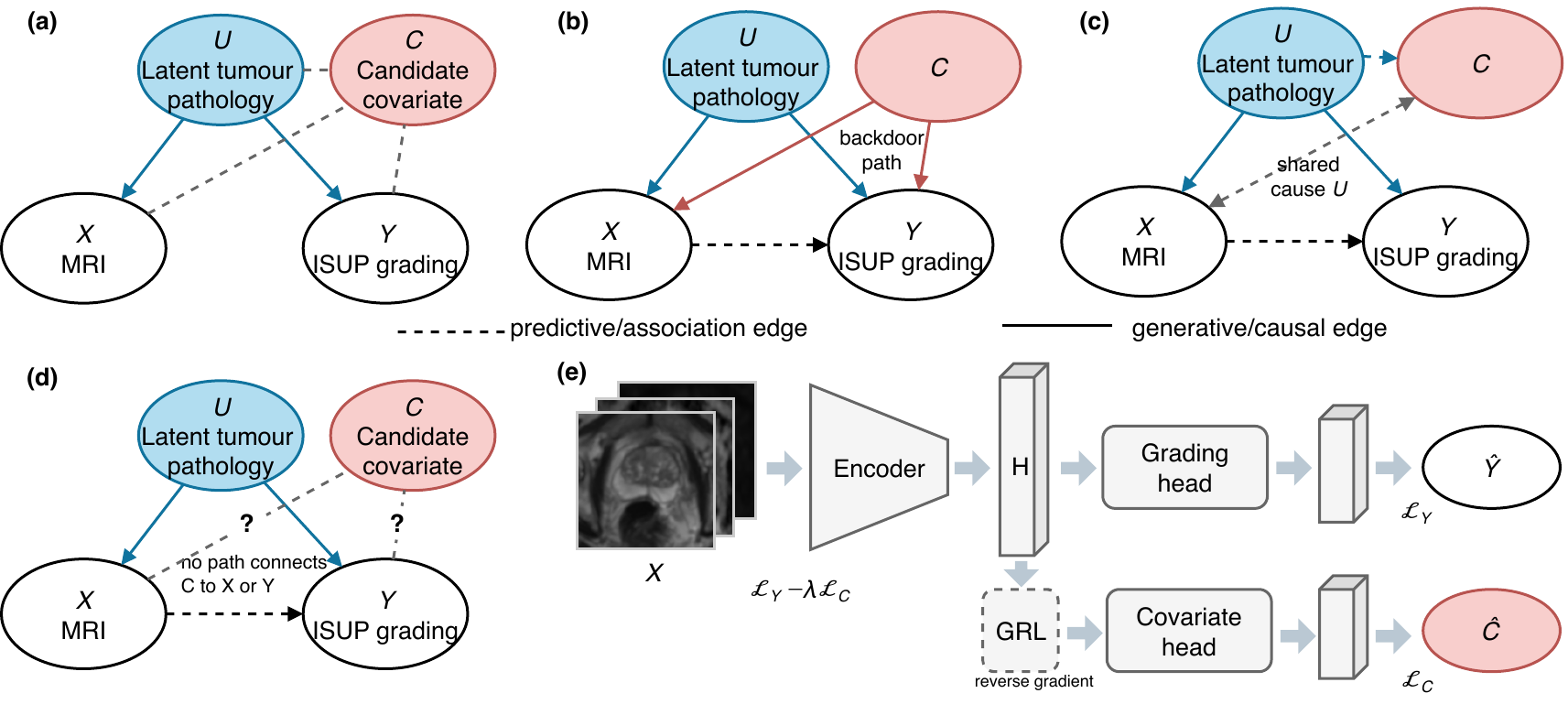}
    \caption{(a) to (d): conceptual hypotheses for candidate covariate \(C\). Blue denotes latent tumour pathology \(U\) and biology-related pathways, red denotes the candidate covariate, and white nodes denote observed MRI \(X\) and ISUP grade \(Y\). 
    The covariate may act as a shortcut/confounder, a disease-related proxy, or an irrelevant covariate. (e): adversarial invariance model for testing candidate \(C\): an encoder maps \(X\) to representation \(H\), a grading head predicts \(\hat{Y}\), and a covariate head predicts \(\hat{C}\) through a gradient reversal layer \cite{ganin2016domain}, reducing \(C\)-decodability while preserving grade-predictive information.
    }
    \label{fig:causal_graph}
\end{figure}

For each candidate covariate, we consider three possible roles, illustrated in Fig.~\ref{fig:causal_graph}. First, \(C\) may behave as a nuisance or confounding candidate if it contributes to a non-pathological association, 
for example through a back-door-type path \(X\leftarrow C\rightarrow Y\). In this case, decodable information about \(C\) may represent a shortcut, and reducing it potentially improves generalisation. Second, \(C\) may be a disease-related proxy if its association with \(X\) or \(Y\) is mediated by, or shares causes with, latent pathology \(U\). Thus, information correlated with \(C\) may also carry grade-relevant signal, so suppressing it could be disadvantageous. Third, \(C\) may be weakly related to the representation or redundant given other MRI features, in which case suppressing it should have little measurable effect.

The causal graph is used only to define representation-level hypotheses. We do not estimate the causal effect of MRI on ISUP grade, infer a complete structural causal model, or perform conventional back-door adjustment by conditioning on subgroups. Instead, we ask whether the model is sensitive to the removal of \(C\)-decodable information from MRI representations. For an evaluation metric \(M\), we compare a baseline MRI-only model with an adversarial model trained to suppress a specific covariate \(C\), \(\Delta_M(C)=M_{\mathrm{adv}(C)}-M_{\mathrm{base}}\).
A positive \(\Delta\) is interpreted as evidence that suppressing \(C\)-related information is beneficial, consistent with nuisance or shortcut behaviour. A negative \(\Delta\) suggests that \(C\)-related information carries useful disease-relevant signal. A near-zero \(\Delta\) suggests that \(C\) is not actively used by the predictor, only weakly encoded or redundant.
These interpretations are treated as sensitivity findings rather than sufficient proof of underlying causal graph.

\subsection{Adversarial invariance for clinical-variable suppression}

To implement the representation-level test, we use an adversarial model \cite{ganin2016domain}. Let \(f_{\theta}\) be the MRI encoder and \(H=f_{\theta}(X)\) its latent representation. The grading head \(g_{\phi}\) predicts the ISUP grade distribution,
\(\hat{Y}=g_{\phi}(H)\), and is trained using the grading loss \(\mathcal{L}_{Y}\). A covariate head \(a_{\psi}\) predicts the selected clinical variable,
\(\hat{C}=a_{\psi}(H)\), using the covariate loss \(\mathcal{L}_{C}\). The covariate head is optimised to recover \(C\), whereas the encoder receives the reversed covariate gradient and is therefore optimised to make \(C\) less decodable from \(H\). The resulting objective is
$
    \min_{\theta,\phi} \max_{\psi}
\mathcal{L}_{Y}\big(g_{\phi}(f_{\theta}(X)),Y\big)
-
\lambda \mathcal{L}_{C}\big(a_{\psi}(f_{\theta}(X)),C\big)$,
where \(\lambda\) controls the strength of the invariance constraint. In practice, this objective can be implemented with a gradient reversal layer between the encoder and adversarial head: the adversary covariate head receives gradients that improve prediction of \(C\), while the encoder weights are updated using both reversed gradients that reduce the decodability of \(C\) from \(H\) and those improve representation learning for task $Y$.
In this work, both ISUP grading $\mathcal{L}_{Y}$ and covariate prediction loss $\mathcal{L}_{C}$ are implemented as cross-entropy losses. The framework can be generalised to other task-appropriate objectives depending on the outcome and covariates.

For each candidate covariate \(C\), we train a separate \(C\)-invariant model and compare it with the baseline MRI-only model. This one-variable-at-a-time design avoids imposing a shared functional interpretation and allows the empirical effect of suppressing each covariate to be measured independently.  
At inference, only MRI is required; the clinical covariate is used only during training to define the adversarial objective.

\begin{table}[tb]
\centering
\caption{All clinical variables tested for the adversarial suppressing experiments. 
}
\label{tab:clinical_variables}
\resizebox{0.9\textwidth}{!}{%
\begin{tabular}{llcc}
\toprule
Variable & Definition & No. classes & No. scans \\
\midrule
Age & Age at MRI: $<60$, 60--70, 70--80, $\geq80$ years & 4 & 2903 \\
BMI & Body-mass index: $<25$, 25--30, $\geq30$ kg/m$^2$ & 3 & 2077 \\
Alcohol use & Alcohol frequency category from clinical records & 7 & 1547 \\
Smoking status & Smoking status category from clinical records & 3 & 1652 \\
Cardiovascular disease & Hypertension or cardiovascular disease & 2 & 1869 \\
Respiratory disease & COPD or asthma & 2 & 1874 \\
Abnormal DRE & Abnormal digital rectal examination & 2 & 467 \\
Diabetes & Recorded diagnosis of diabetes mellitus & 2 & 1938 \\
PI-RADS score & Scan-level PI-RADS score, 1--5 & 5 & 2598 \\
PI-RADS binary cutoff & PI-RADS 4--5 versus PI-RADS 1--3 & 2 & 2598 \\
PSA value & Pre-MRI PSA: $<4$, 4--10, 10--20, $\geq20$ ng/mL & 4 & 1851 \\
Prostate volume & MRI-derived prostate volume: $<30$, 30--60, $\geq60$ mL & 3 & 2646 \\
\bottomrule
\end{tabular}%
}
\end{table}

\section{Experiments and Results}
\noindent\textbf{Data} 
The UCLH clinical-record dataset comprised patients who underwent prostate MRI between April 1, 2019, and March 30, 2024. 
Data were extracted from UCLH Epic data warehouses and linked to prostate MRIs from PACS/VNA, with radiology and histopathology reports obtained from relevant systems.
In total, 2,038 patients with 2,903 pre-treatment prostate MRI scans and corresponding ISUP grades were available. This dataset was split into training, validation and test sets with a ratio of 7:1:2 at the patient level to prevent leakage from repeated scans. 
Clinical covariates were aligned to each MRI scan. 
Pre-imaging clinical measurements, such as PSA, were aligned to the scan date using the closest available value within 3 months before MRI. MRI-derived covariates, including PI-RADS score and prostate volume, were extracted from the corresponding post-MRI radiology report.
A detailed candidate covariate definition is provided in Tab.~\ref{tab:clinical_variables}. 
Continuous variables were binned into clinically interpretable categories to enable a coherent adversarial classification setup and reduce sensitivity to skewed distributions and outliers.
For samples with missing values for a covariate, the adversarial loss was not used, i.e. all samples contributed to the ISUP classification loss, whereas only samples with observed covariate values contributed to training the corresponding adversarial head.
The PROMIS dataset \cite{ahmed2017diagnostic,wang2025promis} comprised MRI from 576 patients with corresponding ISUP grades, acquired between 17 May 2012 and 9 November 2015, and was held out entirely for external validation. 
PROMIS was used only to test models trained on UCLH as it does not contain clinical variables for adversarial training.

All mpMRI scans from both datasets were resampled to a spatial resolution of \(0.5 \times 0.5 \times 1\mathrm{mm}\) and intensity-normalised. DWI and ADC were spatially aligned to T2W images. Data augmentation was performed following ProFound \cite{wang2026profound}.

\noindent\textbf{Comparison experiments}
The primary task was binary classification of ISUP Grade Group \(\geq 2\) versus \(<2\). 
We used ProFound as the main classification encoder backbone, and additionally explored ResNet-18 on the primary binary task.
We also explored ISUP binary cutoff of \(\geq 3\) and \(\geq 4\), and ordinal grade classification. The best checkpoint was selected using validation AUC and quadratic weighted kappa (QWK) for binary and ordinal classification task receptively.

\noindent\textbf{Evaluation metrics}
Binary classification was evaluated using AUC, balanced accuracy, and sensitivity at 80\% specificity (Sens@80Spec). Ordinal classification used QWK and balanced accuracy. For the primary binary task, each model configuration was trained with 6 random seeds, with metrics reported as mean performance with 95\% confidence intervals (CIs) estimated by bootstrap resampling of predictions. For remaining settings, each model was trained with 4 seeds.
For each candidate \(C\), we compared the adversarial and the baseline MRI-only model using the paired metric difference.
Statistical significance of performance changes was assessed using paired bootstrap testing with 10,000 resamples. 

\noindent\textbf{Implementation details}
All models were implemented in PyTorch. The encoder $f_\theta$ used ProFound \cite{wang2026profound}, a ConvNeXt v2 \cite{woo2023convnext} based backbone pre-trained with masked autoencoding \cite{he2022masked} on $4$ prostate MRI datasets comprising 5,000 patients. For classification, we attached a two-layer MLP head and fine-tuned the model for 100 epochs with a batch size of 8 using AdamW, an initial learning rate of \(1\times10^{-4}\) and layer-wise learning-rate decay. The learning rate was linearly warmed up for 5 epochs and then decayed using a half-cycle cosine schedule.
Class-balanced weighted sampling is used for ISUP grade classification. For adversarial training, an auxiliary head was attached through a gradient-reversal layer, with adversarial loss weight \(\lambda=1.0\) and the same training schedule.



\begin{table}[tb]
\centering
\caption{AUC for binary ISUP Grade Group \(\geq2\) vs. \(<2\) classification on UCLH and PROMIS. Values are mean AUC (\%) with 95\% CIs across random seeds. \(\Delta\)AUC denotes the paired change in AUC between each adversarial suppressing model and baseline. Statistical significance is denoted by * \(p<0.05\), ** \(p<0.01\), and *** \(p<0.001\).}
\label{tab:auc}
\resizebox{0.75\textwidth}{!}{%
\begin{tabular}{ll|lllll|llll}
\toprule
 &  & \multicolumn{4}{l}{UCLH} &  & \multicolumn{4}{l}{PROMIS} \\ \cline{1-1} \cline{3-6} \cline{8-11} 
Variable &  & AUC & 95\%CI & $\Delta$AUC &  &  & AUC & 95\%CI & $\Delta$AUC &  \\ \cline{1-1} \cline{3-6} \cline{8-11} 
MRI-only Baseline &  & 67.63 & (67.06, 68.30) &  &  &  & 64.60 & (63.47, 65.45) &  &  \\
Age &  & 68.86 & (68.12, 69.61) & +1.23 & *** &  & 65.38 & (64.58, 66.19) & +0.78 &  \\
BMI &  & 68.47 & (67.83, 69.04) & +0.84 & * &  & 64.10 & (63.42, 65.16) & --0.50 &  \\
Alcohol use &  & 69.04 & (68.36, 69.71) & +1.42 & *** &  & 65.07 & (64.65, 65.54) & +0.47 & * \\
Smoking status &  & 68.36 & (67.58, 68.99) & +0.74 &  &  & 64.33 & (62.95, 65.71) & --0.27 &  \\
Cardiovascular disease &  & 68.23 & (67.43, 68.96) & +0.60 &  &  & 65.33 & (63.62, 67.03) & +0.72 &  \\
Respiratory disease &  & 68.30 & (67.32, 69.27) & +0.67 &  &  & 64.16 & (63.07, 65.02) & --0.44 &  \\
Abnormal DRE &  & 68.12 & (67.11, 69.03) & +0.49 &  &  & 64.33 & (63.09, 65.60) & --0.27 &  \\
Diabetes &  & 68.27 & (67.94, 69.87) & +0.64 &  &  & 64.33 & (61.53, 66.60) & --0.27 &  \\
PI-RADS score &  & 67.85 & (67.15, 68.58) & +0.22 &  &  & 63.24 & (61.78, 64.96) & --1.36 &  \\
PI-RADS binary cutoff &  & 68.21 & (67.19, 69.10) & +0.58 &  &  & 61.80 & (60.67, 62.94) & --2.80 & *** \\
PSA value &  & 65.72 & (64.34, 67.01) & --1.91 & ** &  & 63.06 & (62.12, 64.01) & --1.54 & *** \\
Prostate volume &  & 60.02 & (56.54, 62.35) & --7.61 & *** &  & 61.46 & (60.21, 62.71) & --3.15 & *** \\ \bottomrule
\end{tabular}%
}
\end{table}

\noindent\textbf{Results} Figure~\ref{fig:causal_graph} and Table~\ref{tab:auc} summarise the effect of suppressing each clinical covariate relative to the MRI-only baseline. On the UCLH test set, the baseline AUC was 67.63\% (95\% CI: 67.06--68.30). Enforcing invariance to age, BMI, and alcohol use improved AUC by 1.23\%, 0.84\%, and 1.42\%, respectively, with age and alcohol use also showing consistent improvements in balanced accuracy and Sens@80Spec. Conversely, suppressing PSA and prostate volume reduced AUC by 1.91\% and 7.61\%, respectively, with drops across all metrics, indicating that these variables carried grade-relevant information in the learned representation.

On PROMIS external validation (inference only), the baseline AUC was 64.60\% (95\% CI: 63.47--65.45). 
Effects were less uniform than on UCLH, but several patterns persisted: alcohol-use invariance improved AUC and Sens@80Spec, whereas suppressing binary PI-RADS, PSA value, and prostate volume reduced AUC by 2.80\%, 1.54\%, and 3.15\%, respectively. The PSA-related suppressing effects were attenuated on this external cohort.

Supporting experiments showed broadly consistent trends. Using ResNet-18, prostate volume suppression again caused the largest degradation, reducing AUC by 4.97\% and Sens@80Spec by 5.02\%. Alternative binary thresholds were of higher variability: for ISUP group $\geq3$, PSA suppression remained harmful for AUC ($-1.58\%$), whereas prostate volume was not significant; for ISUP group $\geq4$, the main findings re-emerged, with prostate volume suppression reducing AUC by 2.71\%, balanced accuracy by 6.64\%, and Sens@80Spec by 6.37\%, and PSA suppression reducing AUC by 2.35\% and Sens@80Spec by 4.41\%. Ordinal ISUP prediction was more challenging and less consistent, as expected for multi-class ordinal grading from MRI. Nevertheless, prostate volume remained the most detrimental suppressed variable, reducing QWK by 0.073 and AUC by 7.71\%, with PSA and PI-RADS score also showing concordant decreases. 

\begin{figure}[tb]
    \centering
    \includegraphics[width=\textwidth]{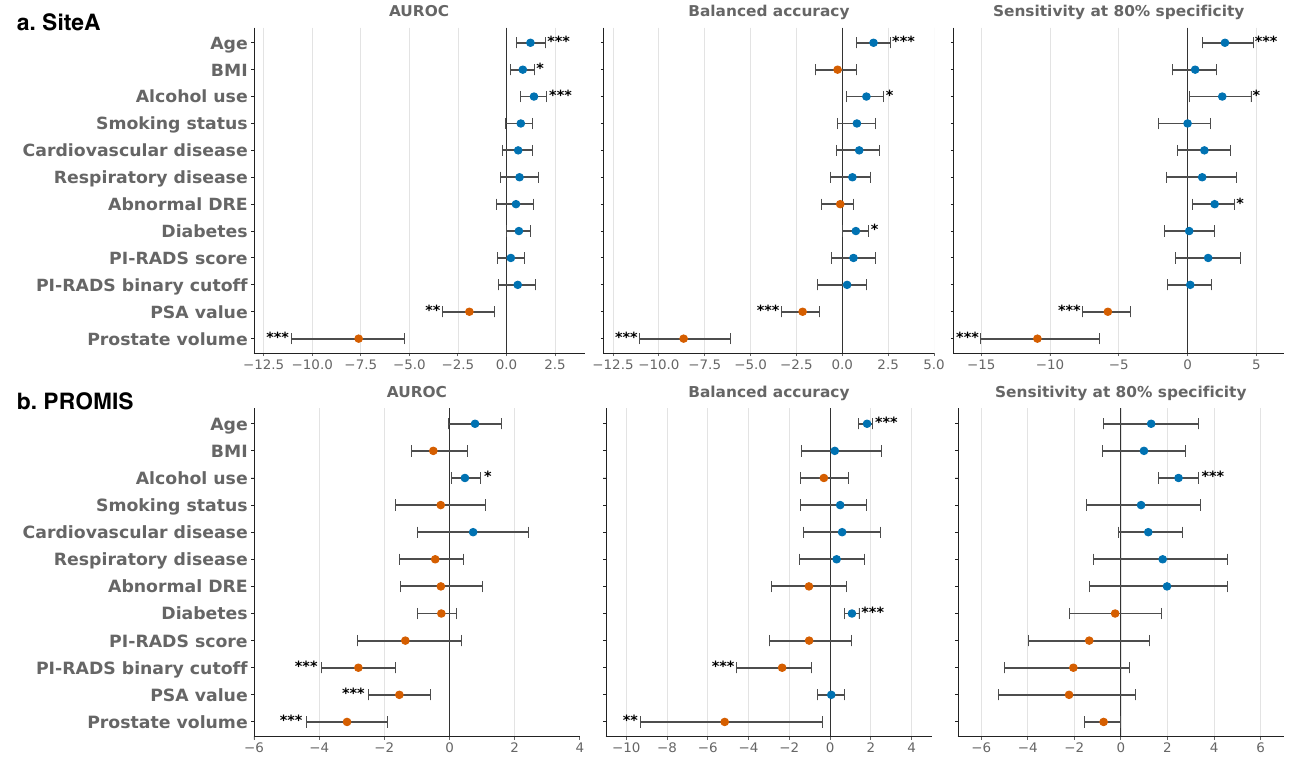}
    \caption{Paired change in binary ISUP grade \(\geq 2\) classification performance after suppressing each clinical covariate on UCLH and PROMIS across AUC, balanced accuracy, and sensitivity at 80\% specificity. Points denote the mean difference between the adversarial suppressing model and the MRI-only baseline, and bars denote 95\% CI. Orange and blue indicates performance decrease and increase relative to baseline, respectively. Statistical significance is denoted by * \(p<0.05\), ** \(p<0.01\), and *** \(p<0.001\).}
    \label{fig:ruleout_forest}
\end{figure}

\section{Discussion and Conclusion}
This study used adversarial covariate suppression to probe whether prostate MRI cancer grading models rely on shared information contained in common clinical variables. In the primary task, suppressing prostate volume caused the largest performance degradation, with PSA suppression also disadvantageous, suggesting task-relevant information encoded in variables. In contrast, suppressing those demographic or lifestyle-related variables, including age, BMI and alcohol use, improved discrimination, suggesting removal of non-generalising information. Similar trends on PROMIS and with ResNet-18 indicate that these effects were not limited to one dataset or backbone, supporting adversarial suppression as a practical tool for distinguishing clinically useful covariate signal from potentially harmful covariate dependence.

It is noteworthy that the limited contribution of PSA in PROMIS may reflect the existing pathway-defined cohort selection. PROMIS was a referred diagnostic cohort with men with clinical suspicion of prostate cancer, commonly including elevated PSA, and therefore may under-represent low-PSA cancers. This referral conditioning could attenuate the incremental value of PSA-related information in external validation, whereas prostate volume may remain more directly encoded in MRI anatomy. Future validation in MRI-naive screening cohorts would be valuable to determine whether PSA-related representation effects differ before PSA-based referral selection, such as those found in screening population.

This work has limitations. Each covariate was tested independently, whereas clinical variables may interact; future work should study covariate combinations, permutations and interactions. Continuous variables were discretised for adversarial classification, and future work could explore raw or log-normalised values with regression heads to assess whether within-bin clinical information remains in the MRI representation. Performance changes after suppression provide candidates for causal follow-up, enabling further studies to formally test whether identified variables satisfy causal definitions of confounding. Although not explored further here, this is also a promising approach to improve cancer detection model generalisability under practical data constraints and high label variability.

In summary, adversarial covariate suppression provides a representation-level approach for assessing covariate dependence in prostate MRI grading models. The divergent effects across variables highlight the need to distinguish useful disease-related signal from less generalisable dependencies.

\bibliographystyle{splncs04}
\bibliography{ref}
%

\end{document}